\documentclass[runningheads]{llncs}

\usepackage{iciap}

\usepackage{iciapabbrv}

\usepackage{graphicx}
\usepackage{booktabs}

\usepackage[accsupp]{axessibility}  

\usepackage{hyperref}

\usepackage{orcidlink}

\usepackage{caption}
\usepackage{subcaption}

\begin{document}

\title{Augmented Reality in Cultural Heritage:\newline A Dual-Model Pipeline for 3D Artwork Reconstruction} 

\titlerunning{Augmented Reality in Cultural Heritage}

\author{Daniele Pannone\inst{1}\orcidlink{0000-0001-6446-6473} \and
Alessia Castronovo \inst{1} \and
Maurizio Mancini\inst{1}\orcidlink{0000-0002-9933-8583} \and \\Gian Luca Foresti\inst{2}\orcidlink{0000-0002-8425-6892} \and Claudio Piciarelli\inst{2}\orcidlink{0000-0001-5305-1520} \and Rossana Gabrieli\inst{1} \and \\Muhammad Yasir Bilal \inst{1}\orcidlink{2222--3333-4444-5555} \and Danilo Avola\inst{1}\orcidlink{0000-0001-9437-6217} }

\authorrunning{Pannone et al.}

\institute{Department of Computer Science, Sapienza University, Italy
\email{\{pannone,castronovo,m.mancini,gabrieli,bilal,avola\}@di.uniroma1.it} \and
Department of Mathematics and Computer Science, University of Udine, Italy
\email{\{gianluca.foresti,claudio.piciarelli\}@uniud.it}\\
}

\maketitle

\begin{abstract}
This paper presents an innovative augmented reality pipeline tailored for museum environments, aimed at recognizing artworks and generating accurate 3D models from single images. By integrating two complementary pre-trained depth estimation models, i.e., GLPN for capturing global scene structure and Depth-Anything for detailed local reconstruction, the proposed approach produces optimized depth maps that effectively represent complex artistic features. These maps are then converted into high-quality point clouds and meshes, enabling the creation of immersive AR experiences. The methodology leverages state-of-the-art neural network architectures and advanced computer vision techniques to overcome challenges posed by irregular contours and variable textures in artworks. Experimental results demonstrate significant improvements in reconstruction accuracy and visual realism, making the system a highly robust tool for museums seeking to enhance visitor engagement through interactive digital content.
\keywords{Deep Learning \and Augmented Reality \and Cultural Heritage}
\end{abstract}

\section{Introduction}
\label{sec:intro}
Advances in computer vision allowed machines to extract meaningful information from visual data, enabling a wide range of applications such as autonomous navigation in robotics \cite{avola2019slam,avola2019visual}, environmental analysis for security \cite{avola2020person,kim2023clip}, and Augmented Reality (AR) \cite{avola2022medicinal,corno2025ar}. The fusion of the latter and 3D modeling has opened new possibilities for enriching cultural and artistic experiences. AR, by overlaying digital content onto the physical world, enhances our perception and interaction with the environment. Unlike Virtual Reality (VR), which immerses users in a completely artificial setting, AR augments the real world with digital information, creating a seamless blend of physical and virtual elements. This technology has found applications across various domains, including healthcare, education, design, and notably, cultural heritage preservation and exhibition.

In the realm of art and culture, AR offers unprecedented opportunities to engage audiences by providing detailed, contextual, and interactive experiences. Museums and galleries, for instance, can leverage AR to offer digital guides that not only display artworks but also provide historical context, detailed views, and even virtual reconstructions of ancient settings or original environments of art pieces. This capability transforms passive observation into dynamic interaction, allowing visitors to explore art in novel and immersive ways. Central to the effectiveness of AR in these contexts is the ability to generate accurate 3D models from 2D images. Traditional methods, such as photogrammetry and Structure from Motion (SfM), have been instrumental in reconstructing 3D structures from multiple images. However, these methods often require multiple views and can be computationally intensive. Recent advancements in computer vision and machine learning, particularly through deep neural networks, have enabled the estimation of depth from single images, making 3D modeling more accessible and efficient.

This paper presents a novel approach to enhancing AR experiences in museum settings by developing a pipeline that recognizes artworks and generates corresponding 3D models from single images. Our method combines two state-of-the-art depth estimation models, GLPN and Depth-Anything, to create detailed depth maps, which are then converted into point clouds and subsequently into 3D meshes using the Poisson surface reconstruction algorithm. These models are integrated into an AR environment using Unity and AR Foundation, allowing for interactive and immersive exploration of artworks. The primary contribution of this work is the development of an optimized pipeline for 3D reconstruction from single images, specifically tailored for artistic subjects. Leveraging the strengths of GLPN and Depth-Anything, our approach produces high-quality depth maps that capture global structure and fine details, essential for realistic 3D renderings. This research provides museums and cultural institutions with a tool for more engaging, interactive experiences, bridging the gap between art and technology. To summarize, the following are the contributions of the paper:
\begin{itemize}
    \item \textbf{Optimized Pipeline for 3D Reconstruction:} Developed a streamlined pipeline for generating 3D models from single images, specifically tailored for artistic subjects, enhancing the accessibility and efficiency of 3D modeling in cultural settings;
    \item \textbf{Integration of Advanced Depth Estimation Models:} Combined GLPN and Depth-Anything models to leverage their complementary strengths, achieving high-quality depth maps that capture both global structure and fine details essential for realistic 3D renderings;
    \item \textbf{Point Cloud and Mesh Generation:} Utilized the Poisson surface reconstruction algorithm to convert depth maps into detailed point clouds and subsequently into 3D meshes, ensuring accurate and visually appealing 3D representations;
    \item \textbf{Enhanced Interactive Cultural Experiences:} Provided a tool for museums and cultural institutions to offer more engaging and interactive experiences, bridging the gap between art and technology by allowing visitors to explore artworks in novel and immersive ways.
\end{itemize}

The rest of the paper is structured as follows. Section \ref{sec:sota} highlights the current state-of-the-art of the techniques used for generating 3D models. Section \ref{sec:method} describes in detail the proposed method. Section \ref{sec:experiments} shows the performed experiments and the obtained results. Finally, Section \ref{sec:conclusion} concludes the paper. 

\section{State of the Art}
\label{sec:sota}
The reconstruction of 3D models from images has been a pivotal area of research in computer vision, driven by the need to create immersive and interactive experiences in fields such as cultural heritage, entertainment, and education. Traditional methods for generating 3D models from images include photogrammetry and Structure from Motion (SfM), each with its own strengths and limitations.

Traditional methods such as photogrammetry and Structure from Motion (SfM) remain foundational. Photogrammetry reconstructs detailed 3D models by aligning multiple images and matching keypoints \cite{remondino2006image}, making it ideal for high-detail capture but reliant on high-quality, multi-view images, often difficult to obtain in constrained settings. SfM, by contrast, estimates both 3D structure and camera poses from image sequences \cite{tron2016sfm}, offering flexibility for dynamic scenes at the cost of high computational demand. Complementary Shape from X techniques \cite{ikeuchi2020light} extract 3D cues from single images using visual properties like shading and silhouette contours, useful in controlled environments. More recently, deep learning has enabled monocular depth estimation, where neural networks infer depth from single images. Models like MiDaS \cite{ranftl2021midas} and GLPN \cite{Kim2022GLPN}, and Depth Anything \cite{depthanything} leverage large datasets to produce accurate depth maps, capturing both global context and fine-grained details, making them especially suitable for artwork digitization. Transformer-based methods have further advanced the field. TripoSR, for example, generates 3D meshes from a single image in under 0.5 seconds \cite{tripoSR2024}, thanks to innovations in model design and training. Similarly, 3D Gaussian Splatting (3DGS) \cite{kerbl3Dgaussians} introduces a novel explicit representation for rapid, high-quality 3D reconstruction, with applications in robotics, mapping, and immersive media. In this context, our work builds upon these established techniques and recent advancements in deep learning to develop an optimized pipeline for 3D reconstruction from single images, specifically tailored for artistic subjects. By leveraging the strengths of both traditional and modern methods, we aim to provide a robust and adaptable solution for significantly enhancing AR experiences in diverse cultural and artistic settings.

\section{Proposed Method}
\label{sec:method}
\begin{figure}[t]
    \centering
    \includegraphics[width=\linewidth]{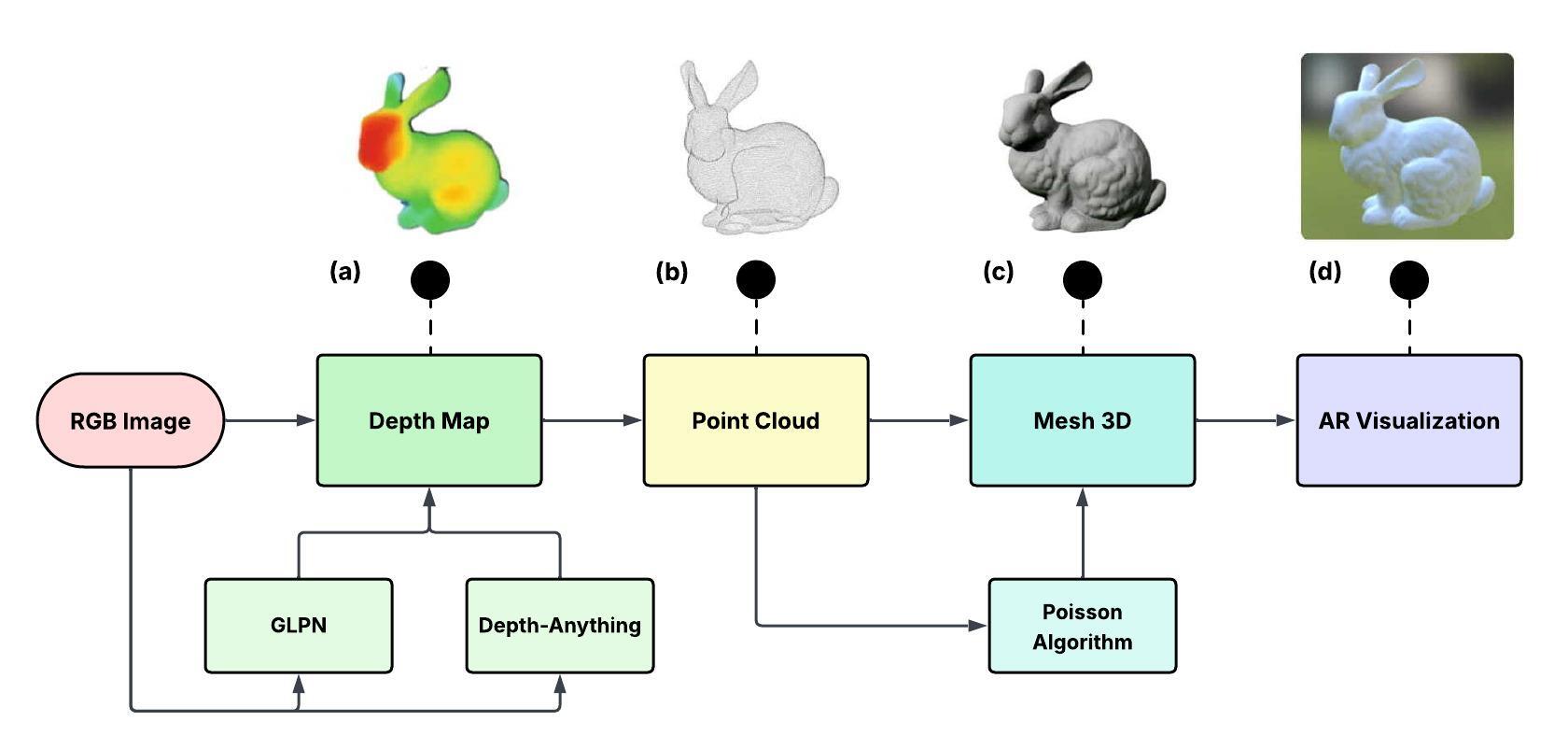}
    \caption{Overview of the proposed pipeline: (a) The input image is processed to generate a depth map using the GLPN and Depth-Anything models. The resulting depth map is used to compute a point cloud (b), which is then converted into a 3D mesh (c) and finally visualized in augmented reality (d).}
    \label{Fig:1}
\end{figure}

In the proposed pipeline, depicted in Fig. \ref{Fig:1}, an input image of an artwork is processed with the aim of generating a 3D mesh, which is then interactively visualized in an augmented reality environment. The following subsections provide a detailed description of each step.

\subsection{Depth Estimation}
\label{subsec:depth_est}
The first step of the pipeline involves predicting a depth map $D \in \mathbb{R}^{1 \times 1 \times h \times w}$ from the input RGB image $A \in \mathbb{R}^{1 \times 3 \times h \times w}$, which is essential for retrieving the three-dimensional information required for the 3D reconstruction. The depth map provides, for each pixel of the image, an estimate of the distance between the corresponding surface point and the camera. Given the absence of depth sensors or stereo vision systems capable of providing metric data, relative depth maps were adopted. In such representations, values are expressed on an arbitrary scale and indicate only which points are closer or farther, without providing absolute measurements in physical units. To generate the depth maps, two pre-trained deep learning models were employed: GLPN and Depth-Anything V2. The GLPN model, presented in \cite{Kim2022GLPN} adopts an architecture that combines a global analysis of the entire image with a local analysis for detail extraction, in order to improve the accuracy of the generated depth map. The architecture includes a Transformer-based encoder to capture global dependencies, along with a decoder that, through the Selective Feature Fusion module, identifies the most relevant features for depth estimation. The model was trained on the NYU Depth V2 dataset, which contains images of indoor scenes, using a data augmentation technique known as Vertical CutDepth. This method preserves the vertical positioning of objects in the image, allowing the model to better distinguish depth layers, such as separating the floor from the walls. Depth-Anything V2 \cite{depthanything} is an enhanced version of the original Depth-Anything model. It leverages the DPT (Dense Prediction Transformer) architecture with DINOv2 backbone \cite{oquab2024dinov2learningrobustvisual}. The model was fine-tuned on large-scale public datasets, such as NYUv2 \cite{Silberman:ECCV12} and KITTI \cite{Geiger2013IJRR}, which contain images annotated with metric depth maps. This supervised training enabled the model to generate highly accurate depth estimations.

The two models were selected for their complementary properties: GLPN excels in distinguishing depth layers, while Depth-Anything offers greater precision in fine details. For this reason, a weighted average of the depth maps predicted by the two models was computed to generate a final map that is accurate and well-balanced for our purpose.

\subsection{Depth Maps Fusion}
\label{subsec:depth_fusion}
Before being processed by the depth estimation models, input images are resized so that their dimensions are multiple of 32, ensuring compatibility with the neural networks employed. During testing, it was observed that using either model in isolation did not yield depth maps of sufficient accuracy for generating precise 3D meshes. Specifically, GLPN proved effective in distinguishing foreground from background but lacked precision in local details, resulting, for example, in flattened facial regions. In contrast, Depth-Anything demonstrated high sensitivity to fine details, such as facial features or fingertips, but produced disproportionate reconstructions due to inconsistent depth scaling. To leverage the strengths of both models, a pixel-wise weighted fusion strategy was employed. Let $D_{GLPN} \in \mathbb{R}^{H \times W}$ and $D_{DA} \in \mathbb{R}^{H' \times W'}$ denote the depth maps produced by GLPN and Depth-Anything, respectively. As the outputs may differ in spatial resolution, $D_{DA}$ was first resized to the resolution of $D_{GLPN}$ using bilinear interpolation:
\begin{equation}
\tilde{D}_{DA} = Resize(D_{DA},H,W).
\end{equation}

The final combined depth map $D_{combined} \in \mathbb{R}^{H \times W}$ was computed via a convex combination:
\begin{equation}
\label{eq:depth_fusion}
D_{combined}(x, y) = \alpha \cdot D_{GLPN}(x, y) + (1 - \alpha) \cdot \tilde{D}_{DA}(x, y),
\end{equation}
\noindent where $\alpha = 0.97$ was empirically determined to preserve the global structural consistency of GLPN while incorporating local detail enhancements from Depth-Anything. Notably, even a small contribution from Depth-Anything $1 - \alpha = 0.03$ was found to significantly improve the perceptual fidelity of the final depth map. Despite this fusion, the resulting depth map still exhibited inconsistencies due to the heterogeneous scale calibration of the two models. This manifested in depth inversion and stretching artifacts, with nearby surfaces appearing erroneously distant, and vice versa. To mitigate these effects, a range normalization was applied to scale depth values into a target interval $[d_{\min}, d_{\max}] = [0.6, 1.0]$. The normalization was performed as:
\begin{equation}
\label{eq:depth_normalization}
D_{norm}(x, y) = d_{\min} + \left( \frac{D_{combined}(x, y) - \min(D_{combined})}{\max(D_{combined}) - \min(D_{combined})} \right) \cdot (d_{\max} - d_{\min}).
\end{equation}
\noindent This operation ensured consistency in-depth scale across samples, aligning the depth values with the expected spatial bounds of the scene. The normalized depth map $\mathbf{D}_{norm} \in \mathbb{R}^{H \times W}$ was then converted into a 3D point cloud $\mathcal{P} = {\mathbf{p}_i}_{i=1}^{N}$, where each point $\mathbf{p}_i = (x_i, y_i, z_i) \in \mathbb{R}^3$ was computed through standard back-projection using the camera intrinsic matrix $\mathbf{K} \in \mathbb{R}^{3 \times 3}$. Let $(u_i, v_i)$ denote the pixel coordinates in the image plane. The corresponding 3D point is given by:
\begin{equation}
\label{eq:back_projection}
p_i = K^{-1} \cdot 
\begin{bmatrix}
u_i \\
v_i \\
D_{norm}(u_i, v_i)
\end{bmatrix}.
\end{equation}

After generating the raw point cloud $\mathcal{P}$, a statistical outlier removal procedure was applied to eliminate noisy points. Subsequently, surface normals were estimated for each point $\mathbf{p}_i$, a critical step for the application of the Poisson Surface Reconstruction algorithm, which was used to generate a watertight mesh from the point cloud. The resulting mesh was post-processed in Blender, a 3D modeling and rendering software. This phase included manual segmentation to remove the background and retain only the main subject of the scene. Furthermore, material properties were assigned to the mesh to enable correct rendering of vertex colors in the Unity game engine.

\subsection{3D Reconstruction}
\label{subsec:3d_rec}
The process begins with the estimation of a depth map $ D: \Omega \subset \mathbb{R}^2 \to \mathbb{R}^+ $, where $ \Omega $ denotes the image domain. This depth map is combined with the input RGB image $ I: \Omega \to \mathbb{R}^3 $, resulting in an RGB-D image $ \mathcal{I}_{RGB-D}(u,v) = (I(u,v), D(u,v)) $, where each pixel $ (u, v) \in \Omega $ now encodes both color and depth information. This RGB-D representation is crucial for constructing an initial 3D model of the scene. To recover the spatial structure of the scene, the RGB-D image is projected into 3D space using the Pinhole camera model. Each pixel $ (u,v) $ in the image, with associated depth value $ D(u,v) $, is mapped to a 3D point $ \mathbf{P}(u,v) = (X, Y, Z) \in \mathbb{R}^3 $ in the camera coordinate system as follows:
\begin{equation}
	\label{eq:1}
	X = \frac{(u - c_x) \cdot D(u,v)}{f_x}, \quad 
	Y = \frac{(v - c_y) \cdot D(u,v)}{f_y}, \quad 
	Z = D(u,v),
\end{equation}

\noindent where $ (c_x, c_y) $ denote the principal point (optical center) of the camera, and $ f_x, f_y $ represent the focal lengths along the horizontal and vertical axes, respectively. This mapping yields a point cloud $ \mathcal{P} = \{ \mathbf{P}_i \}_{i=1}^{N} \subset \mathbb{R}^3 $, where each point corresponds to a pixel in the RGB-D image. While the point cloud provides a discrete sampling of the object surface, it lacks topological structure. To reconstruct a continuous surface, the Poisson surface reconstruction algorithm \cite{Kazhdan2006Poisson} is applied. This method interprets the set of oriented points $ \{ (\mathbf{P}_i, \mathbf{n}_i) \} $, where $ \mathbf{n}_i \in \mathbb{S}^2 $ denotes the surface normal at point $ \mathbf{P}_i $, as samples of an implicit surface defined by the gradient of an indicator function $\chi: \mathbb{R}^3 \to \{0, 1\} $, where $\chi(\mathbf{x}) = 1$ if $x$ lies inside the object, $\chi(\mathbf{x}) = 0$ otherwise. The goal is to recover a function $ \chi $ such that its gradient $ \nabla \chi $ approximates the input vector field $ \mathbf{V}(\mathbf{x}) $, built from the normal vectors $ \mathbf{n}_i $. The reconstruction task is formulated as a Poisson equation:
\begin{equation}
	\label{eq:2}
	\Delta \chi = \nabla \cdot \mathbf{V},
\end{equation}
\noindent where $ \Delta $ denotes the Laplace operator, and $ \nabla \cdot \mathbf{V} $ is the divergence of the vector field $ \mathbf{V} $. Solving this equation yields a scalar field $ \chi $ whose zero level set or isosurface (typically at $\chi = 0.5$) represents the reconstructed surface of the object. This formulation provides a principled approach to surface reconstruction that naturally incorporates the orientation and spatial coherence of the data. Moreover, it yields a smooth mesh even in the presence of noise or missing regions in the original point cloud, making it particularly suitable for real-world 3D reconstruction tasks.

\subsection{Augmented Reality Visualization}
\label{subsec:ar_visual}
The final stage of the proposed pipeline consists in the interactive visualization of the 3D mesh through augmented reality. To achieve this, the mesh is imported into the Unity engine, selected for its advanced rendering capabilities and for its support of the ARFoundation framework, which enables cross-platform AR application development, supporting Android devices via ARCore and iOS devices via ARKit. Within Unity, the mesh is configured as a prefab, a reusable asset that can be dynamically instantiated during runtime. To automatically trigger and position the correct mesh in the AR scene, Unity image tracking system, known as AR Tracked Image Manager, was adopted. This component allows the device to recognize images in the real-world environment. These images were included in a reference image library linked to the tracking system. When one of them is detected by the device’s camera, the corresponding prefab is automatically instantiated, spatially anchored to the recognized marker, and displayed in the correct position. This setup enables consistent and accurate visualization of the 3D model within the real-world context.

\section{Experimental Results}
\label{sec:experiments}
This section presents both the implementation choices that influenced the final output and the results obtained from testing the pipeline on various categories of artworks. In particular, Sect. \ref{subsec:implementation} describes the the used tool and dataset for accomplish the proposed task. Sect. \ref{subsec:clip} reports a quantitative evaluation of the generated meshes, based on CLIP similarity scores, to assess the semantic-visual consistency between the 3D output and the original artwork. Finally, Sect. \ref{subsec:qual_analysis} reports a qualitative analysis of the reconstruction outcomes, highlighting the effectiveness and limitations of the pipeline across different types of input images.

\subsection{Dataset and Implementation Details}
\label{subsec:implementation}
The proposed pipeline was implemented using the Python programming language, leveraging the PyTorch library for depth estimation inference and the Open3D library for point cloud and mesh construction. The dataset used for the test was the WikiArt dataset \cite{artgan2018}, together with manually downloaded images of artists and paintings not present in such dataset.
The interactive augmented reality visualization was developed in Unity using the C\# programming language. All tests were performed on Android devices. 

\subsection{CLIP-Based Semantic Consistency}
\label{subsec:clip}
\begin{table}[t]
    \centering
    \caption{CLIP similarity values obtained for the chosen artworks by comparing each artwork with the corresponding mesh generated using Depth-Anything, GLPN, and the proposed pipeline.}
    \begin{tabular}{lccc}
    \toprule
    \textbf{Artwork} \quad & \textbf{Depth-Anything} \quad & \textbf{GLPN} \quad & \textbf{Proposed Pipeline} \\
    \midrule
    Lady with an Ermine          & 0.7040 & 0.5717 & 0.7314 \\
    The Starry Night             & 0.5672 & 0.7025 & 0.7168 \\
    Mona Lisa                    & 0.5732 & 0.5891 & 0.7686 \\
    Relativity                   & 0.5774 & 0.6011 & 0.6928 \\
    Girl with a Pearl Earring    & 0.5048 & 0.6758 & 0.7376 \\
    \bottomrule
    \end{tabular}
    \label{tab:1}
\end{table}
In the absence of ground truth data for the generated 3D meshes, a direct geometric evaluation was not feasible. Therefore, a quantitative evaluation was conducted based on the computation of semantic-visual similarity between the reconstructed meshes and their corresponding original artworks. The evaluation relied on the CLIP model \cite{radford2021learningtransferablevisualmodels}, specifically the ViT-B/32-QuickGELU variant, as implemented in the open clip library \cite{cherti2023reproducible}. The CLIP model is designed to project both images and texts into a shared multimodal embedding space $\mathcal{E} \subset \mathbb{R}^{d}$, where semantically and visually similar inputs are mapped to vectors with small angular distances. Let $f: \mathcal{I} \to \mathcal{E}$ denote the CLIP image encoder, which maps an image $\mathbf{I}$ to a normalized embedding vector $\mathbf{v} = f(\mathbf{I}) \in \mathcal{E}$, such that $\|\mathbf{v}\|_2 = 1$. Given two images, i.e. an original artwork $\mathbf{I}_{art}$ and a rendered image $\mathbf{I}_{mesh}$ of the corresponding reconstructed 3D mesh, their semantic-visual similarity is computed using the cosine similarity metric:
\begin{equation}
\label{eq:cosine_similarity}
sim (\mathbf{I}_{art}, \mathbf{I}_{mesh}) = \frac{f(\mathbf{I}_{art}) \cdot f(\mathbf{I}_{mesh})}{\|f(\mathbf{I}_{art})\|_2 \cdot \|f(\mathbf{I}_{mesh})\|_2}.
\end{equation}
\noindent Since all embedding vectors are $\ell_2$-normalized, Equation \ref{eq:cosine_similarity} simplifies to the dot product of the two embeddings:
\begin{equation}
\label{eq:normalized_similarity}
sim(\mathbf{I}_{art}, \mathbf{I}_{mesh}) = f(\mathbf{I}_{art})^\top \cdot f(\mathbf{I}_{mesh}).
\end{equation}
\noindent The resulting similarity score $sim(\cdot, \cdot) \in [-1, 1]$ quantifies the semantic alignment between the visual content of the original artwork and the reconstructed mesh. A value close to 1 indicates strong semantic similarity, a value near 0 suggests weak or no correlation, and negative values imply semantic opposition. For this evaluation, each original artwork image $\mathbf{I}_{art}^{(i)}$ was compared against a rendered view $\mathbf{I}_{\text{mesh}}^{(i)}$ of its corresponding 3D reconstruction. The reconstructions were obtained using three different approaches: the GLPN-based depth estimation, the Depth-Anything-based depth estimation, and the Proposed fusion-based pipeline. The rendered mesh images $\mathbf{I}_{mesh}^{(i)}$ were captured from a partially lateral viewpoint, as opposed to a strictly frontal perspective, to assess whether the reconstructed 3D geometry retained enough structural and visual cues to be semantically matched with the original 2D artwork. The complete evaluation process can be summarized as follows. Let $N$ denote the number of artwork-mesh pairs. The mean similarity score for each reconstruction method is computed as:

\begin{equation}
\label{eq:mean_similarity}
\overline{sim} = \frac{1}{N} \sum_{i=1}^{N} sim\left( \mathbf{I}_{art}^{(i)}, \mathbf{I}_{mesh}^{(i)} \right).
\end{equation}

\noindent This metric serves as an aggregate indicator of the semantic preservation achieved by each reconstruction pipeline, as judged by the CLIP model’s high-level embedding space. Table \ref{tab:1} reports the similarity values obtained for each variant. The data refer to the same artworks shown in the qualitative evaluation presented in Fig. \ref{Fig:2}, allowing for a direct comparison between visual perception and automated measurement. As can be observed, the proposed pipeline consistently yields higher similarity scores, closer to 1, indicating a stronger perceptual consistency compared to solutions based on individual depth estimation models.
%
\newlength{\tempwidth}
\newcommand{\columnname}[1]
{\makebox[\tempwidth][c]{\textbf{#1}}}

\begin{figure}[t]
\setlength{\tempwidth}{0.15\linewidth}
    \centering
    \columnname{Input} \hfill
    \columnname{Depth-Anything} \hfill
    \columnname{GLPN} \hfill
    \columnname{Ours} \\
    \begin{subfigure}[b]{0.1\textwidth}
         \centering
         \includegraphics[width=\textwidth]{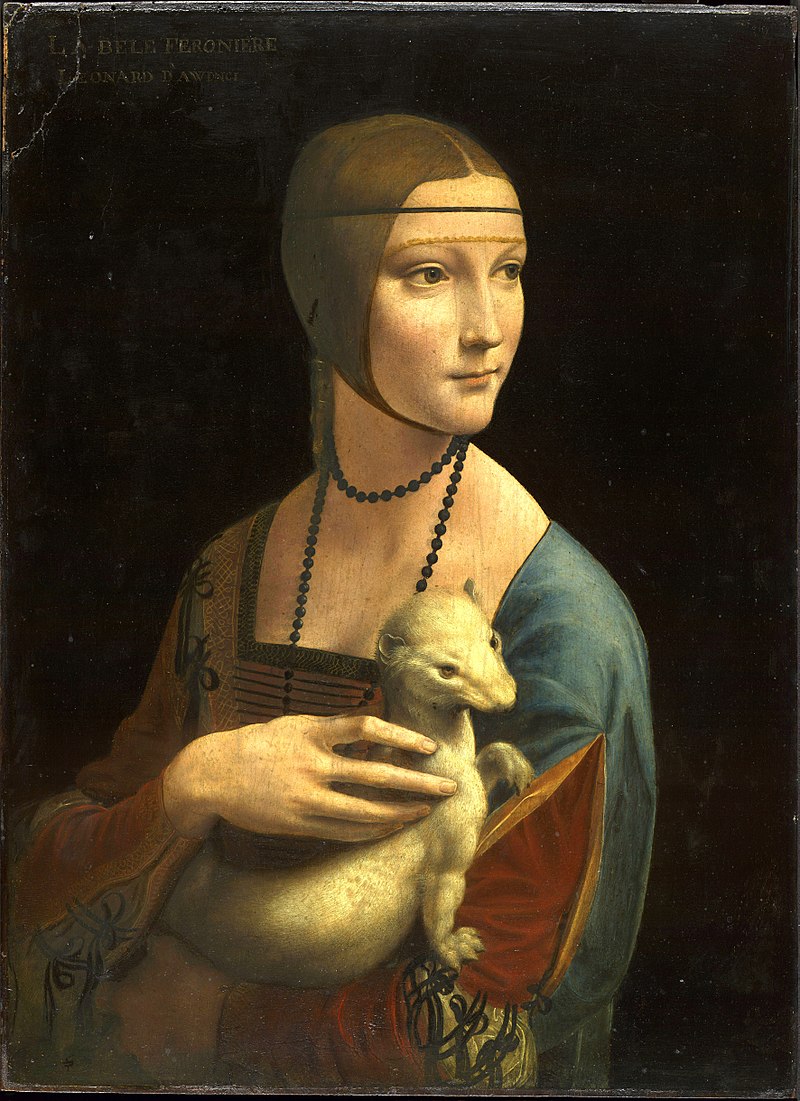}
         \caption{}
     \end{subfigure}
     \hfill
     \begin{subfigure}[b]{0.13\textwidth}
         \centering
         \includegraphics[width=\textwidth]{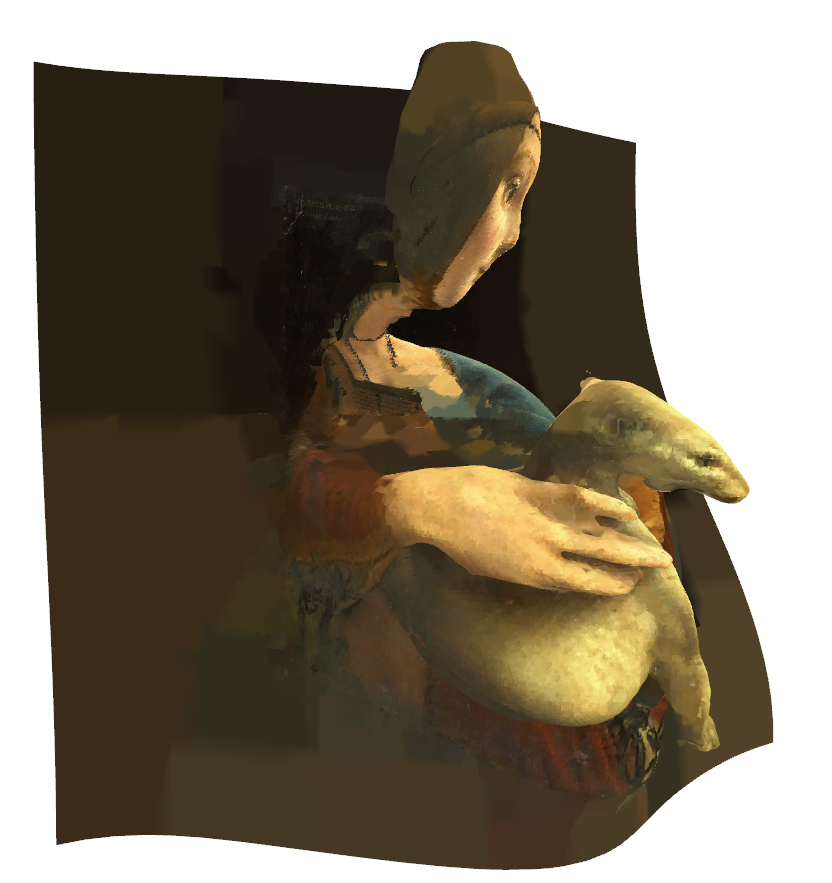}
         \caption{}
     \end{subfigure}
     \hfill
     \begin{subfigure}[b]{0.11\textwidth}
         \centering
         \includegraphics[width=\textwidth]{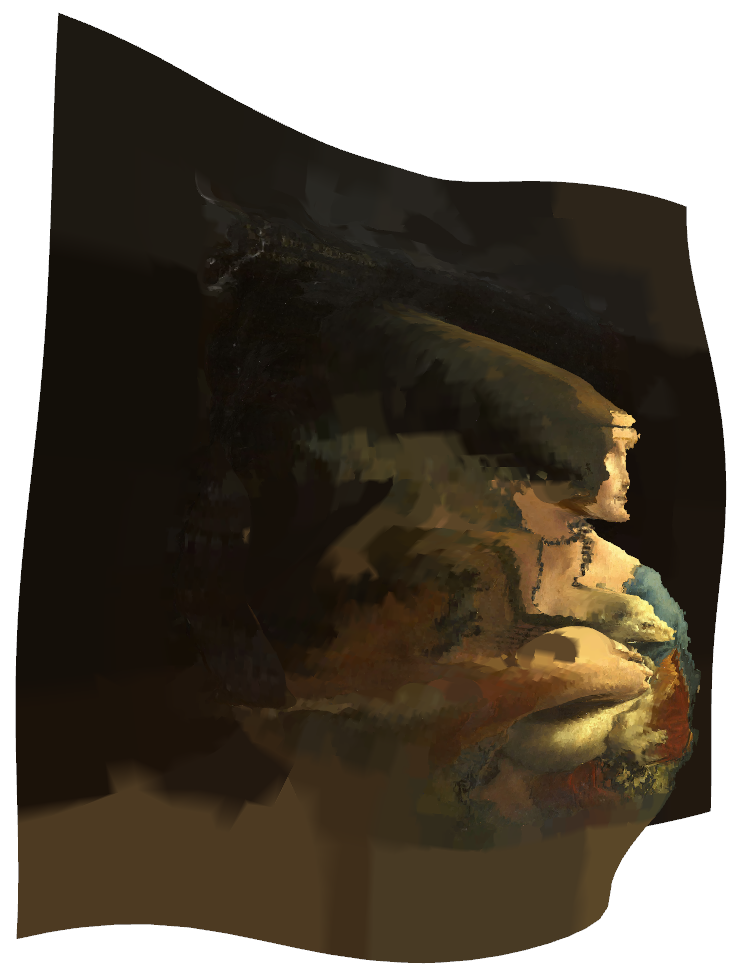}
         \caption{}
     \end{subfigure}
     \hfill
     \begin{subfigure}[b]{0.15\textwidth}
         \centering
         \includegraphics[width=\textwidth]{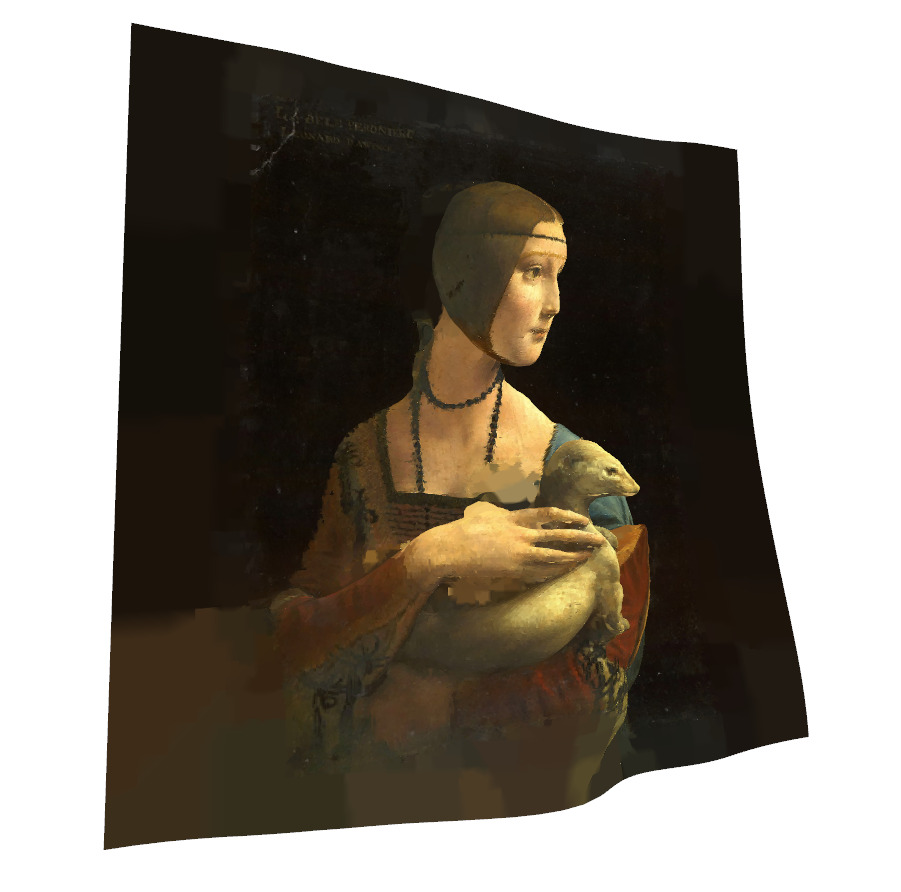}
         \caption{}
     \end{subfigure}

     \begin{subfigure}[b]{0.13\textwidth}
         \centering
         \includegraphics[width=\textwidth]{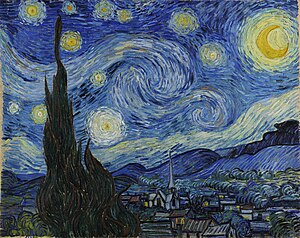}
         \caption{}
     \end{subfigure}
     \hfill
     \begin{subfigure}[b]{0.13\textwidth}
         \centering
         \includegraphics[width=\textwidth]{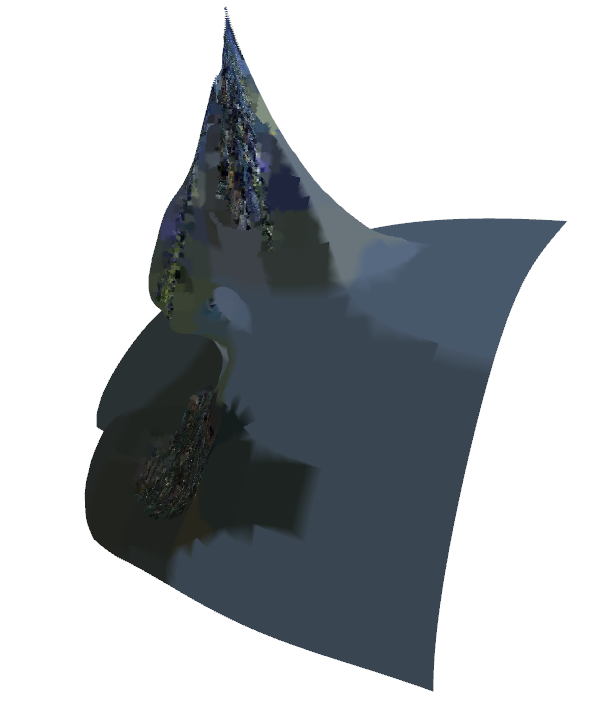}
         \caption{}
     \end{subfigure}
     \hfill
     \begin{subfigure}[b]{0.14\textwidth}
         \centering
         \includegraphics[width=\textwidth]{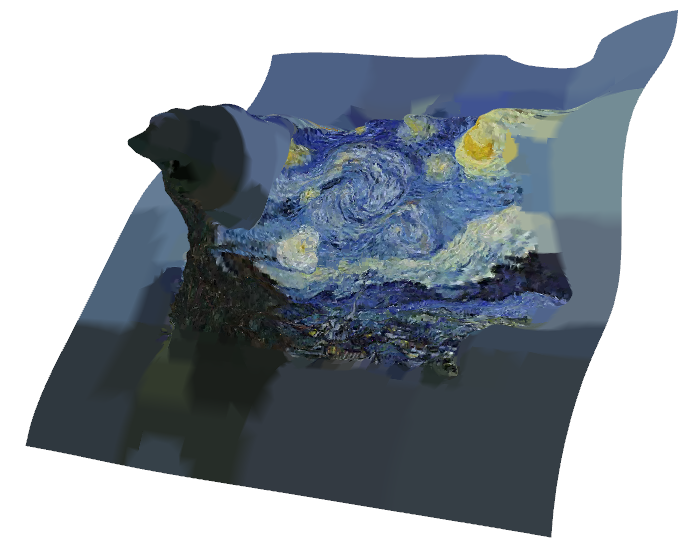}
         \caption{}
     \end{subfigure}
     \hfill
     \begin{subfigure}[b]{0.13\textwidth}
         \centering
         \includegraphics[width=\textwidth]{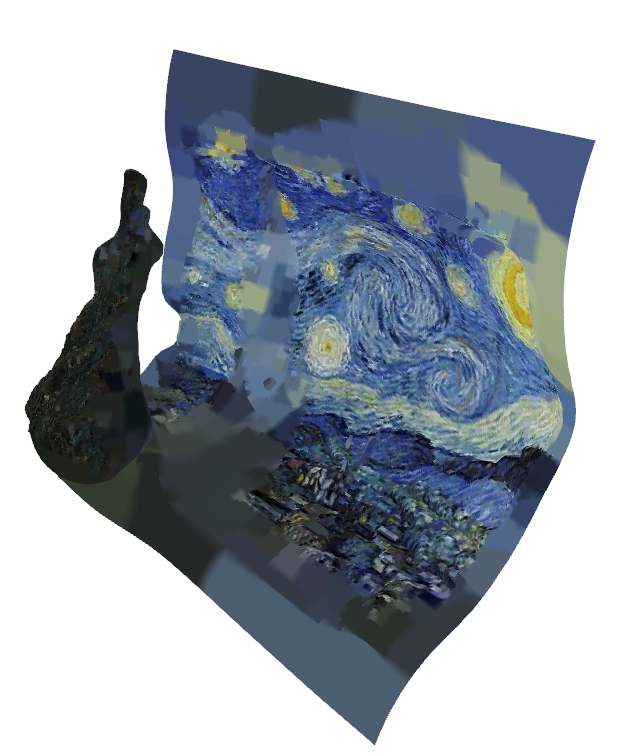}
         \caption{}
     \end{subfigure}

     \begin{subfigure}[b]{0.11\textwidth}
         \centering
         \includegraphics[width=\textwidth]{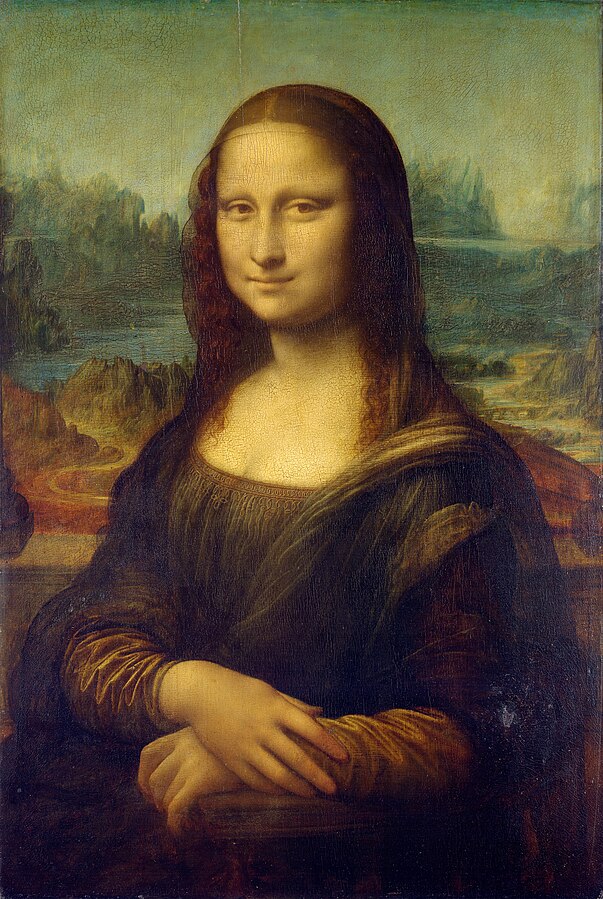}
         \caption{}
     \end{subfigure}
     \hfill
     \begin{subfigure}[b]{0.18\textwidth}
         \centering
         \includegraphics[width=\textwidth]{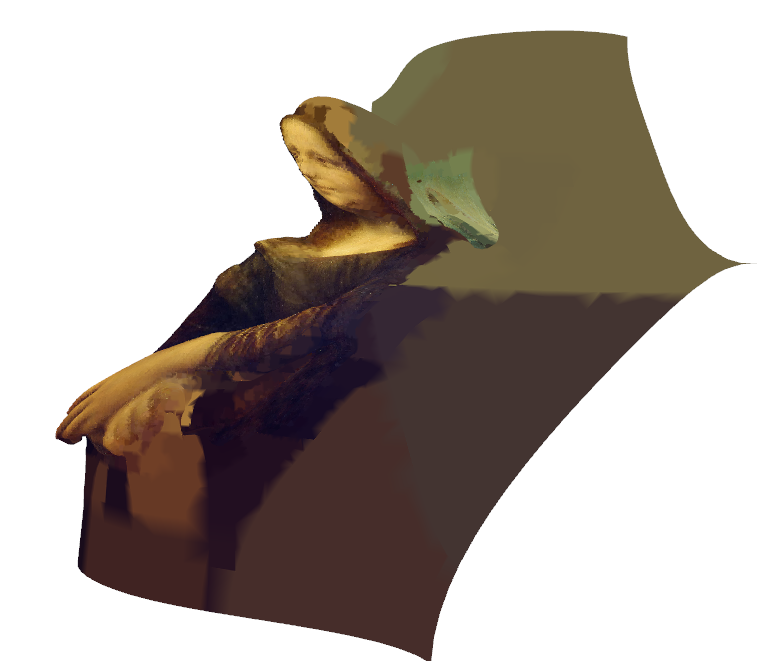}
         \caption{}
     \end{subfigure}
     \hfill
     \begin{subfigure}[b]{0.15\textwidth}
         \centering
         \includegraphics[width=\textwidth]{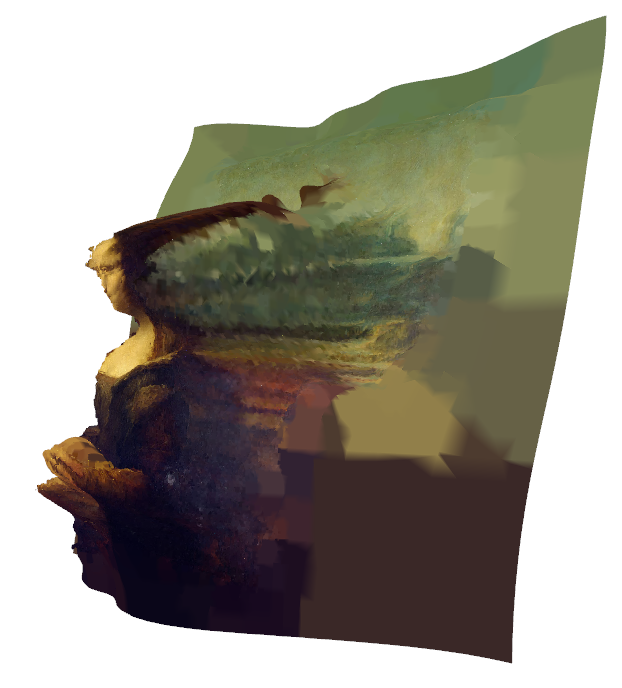}
         \caption{}
     \end{subfigure}
     \hfill
     \begin{subfigure}[b]{0.15\textwidth}
         \centering
         \includegraphics[width=\textwidth]{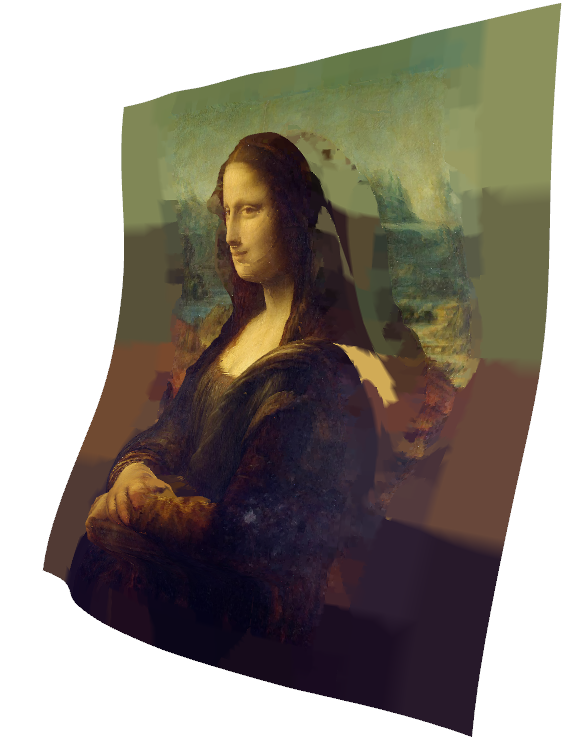}
         \caption{}
     \end{subfigure}

     \begin{subfigure}[b]{0.13\textwidth}
         \centering
         \includegraphics[width=\textwidth]{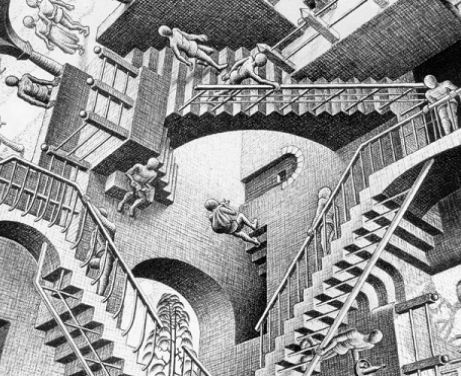}
         \caption{}
     \end{subfigure}
     \hfill
     \begin{subfigure}[b]{0.15\textwidth}
         \centering
         \includegraphics[width=\textwidth]{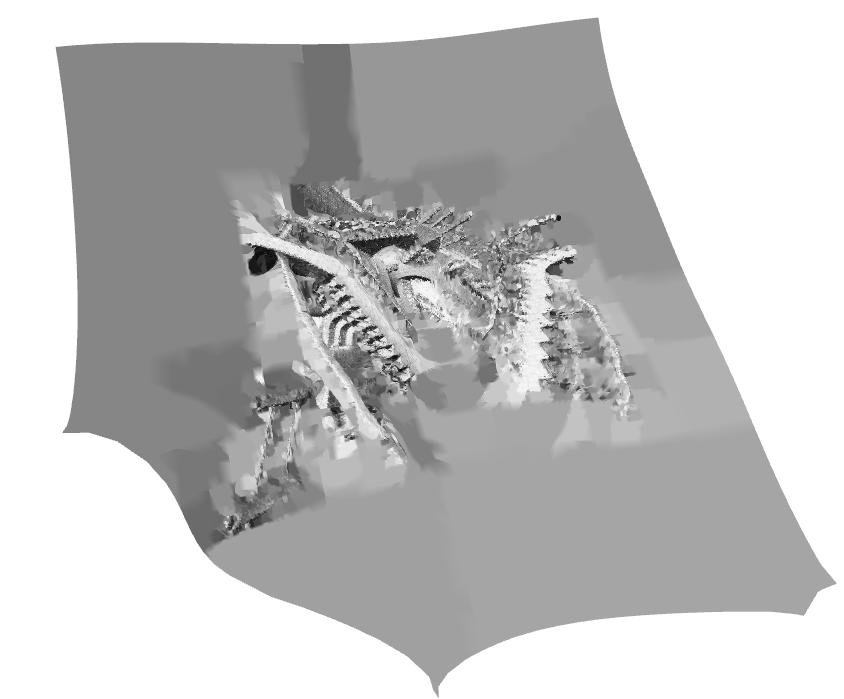}
         \caption{}
     \end{subfigure}
     \hfill
     \begin{subfigure}[b]{0.15\textwidth}
         \centering
         \includegraphics[width=\textwidth]{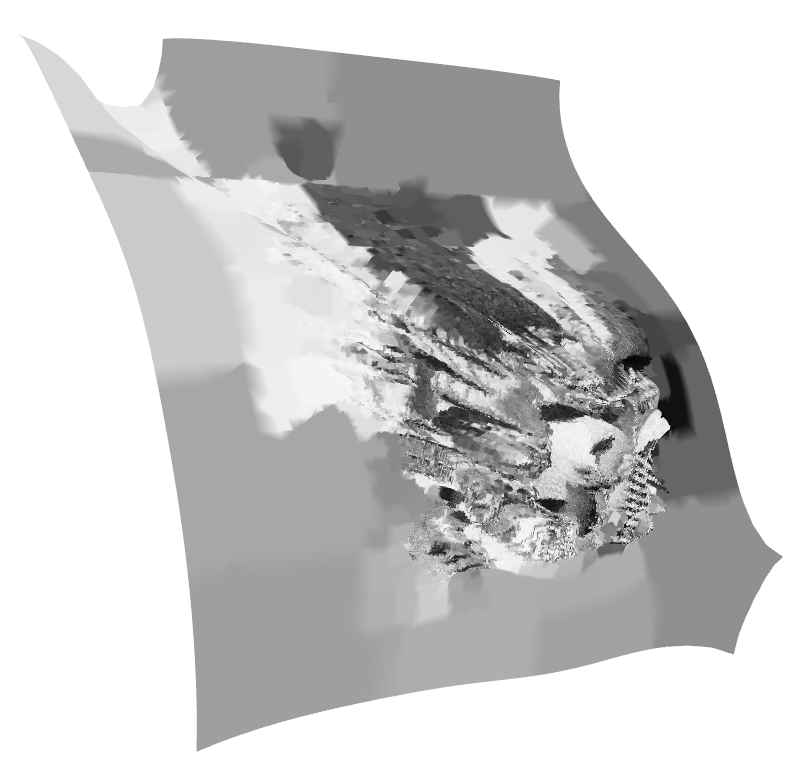}
         \caption{}
     \end{subfigure}
     \hfill
     \begin{subfigure}[b]{0.15\textwidth}
         \centering
         \includegraphics[width=\textwidth]{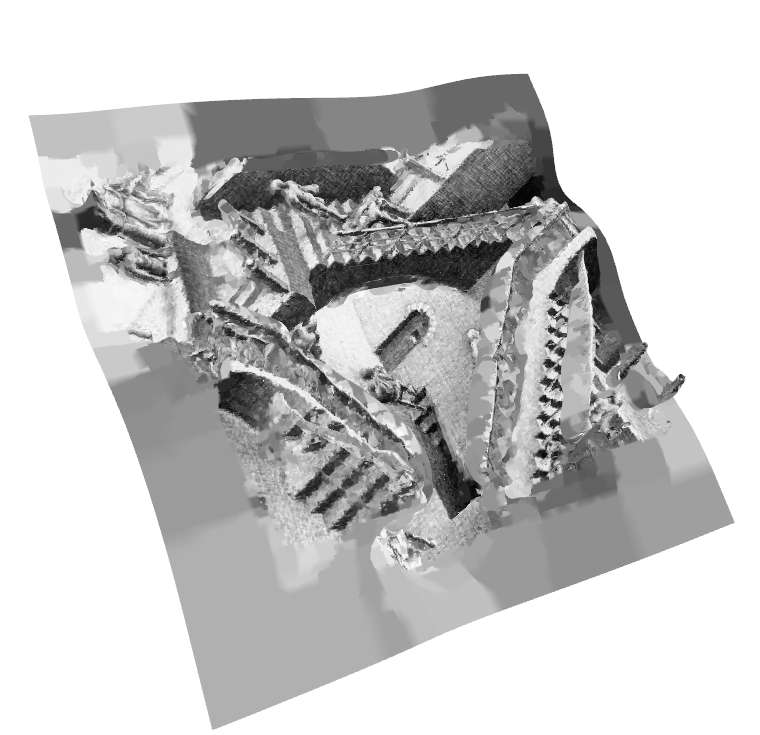}
         \caption{}
     \end{subfigure}

     \begin{subfigure}[b]{0.13\textwidth}
         \centering
         \includegraphics[width=\textwidth]{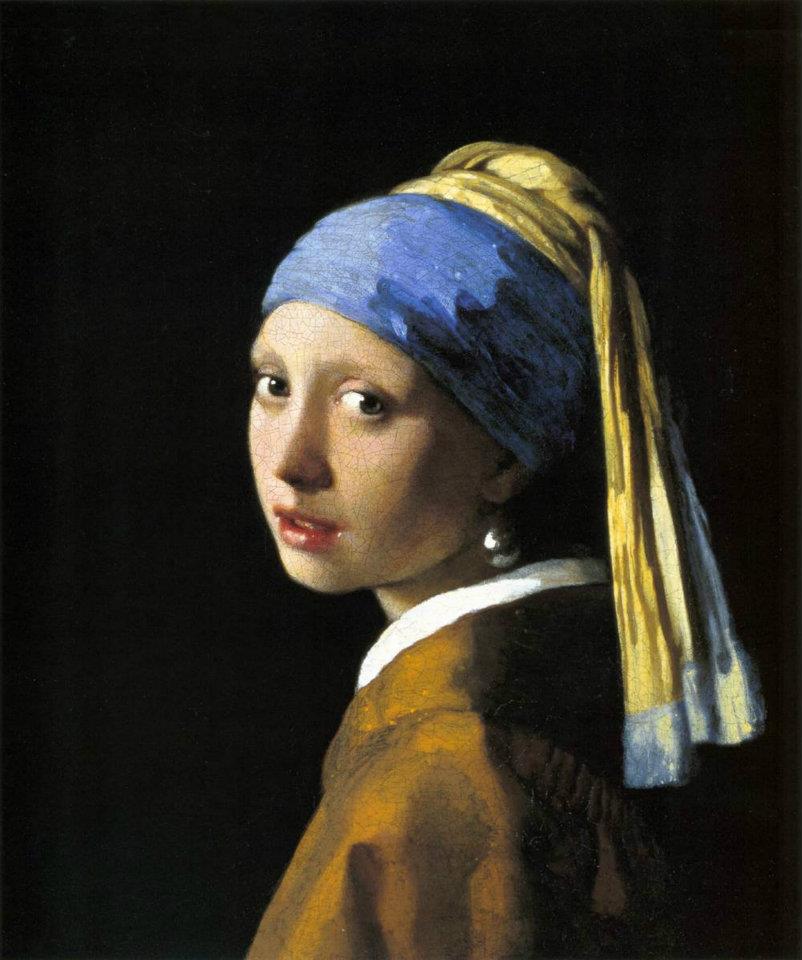}
         \caption{}
     \end{subfigure}
     \hfill
     \begin{subfigure}[b]{0.13\textwidth}
         \centering
         \includegraphics[width=\textwidth]{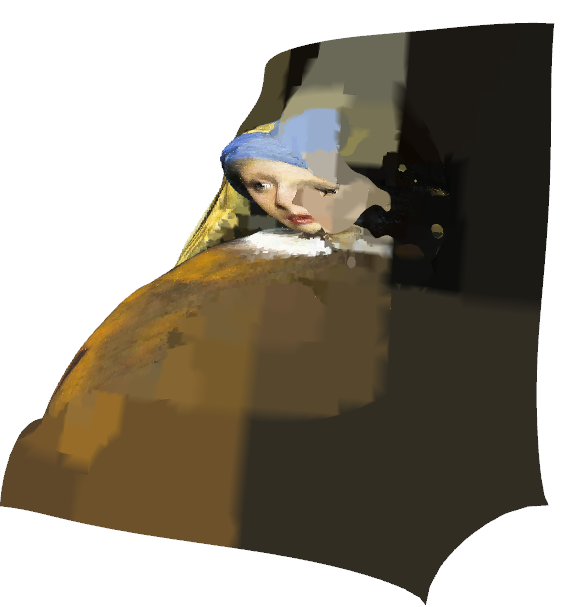}
         \caption{}
     \end{subfigure}
     \hfill
     \begin{subfigure}[b]{0.13\textwidth}
         \centering
         \includegraphics[width=\textwidth]{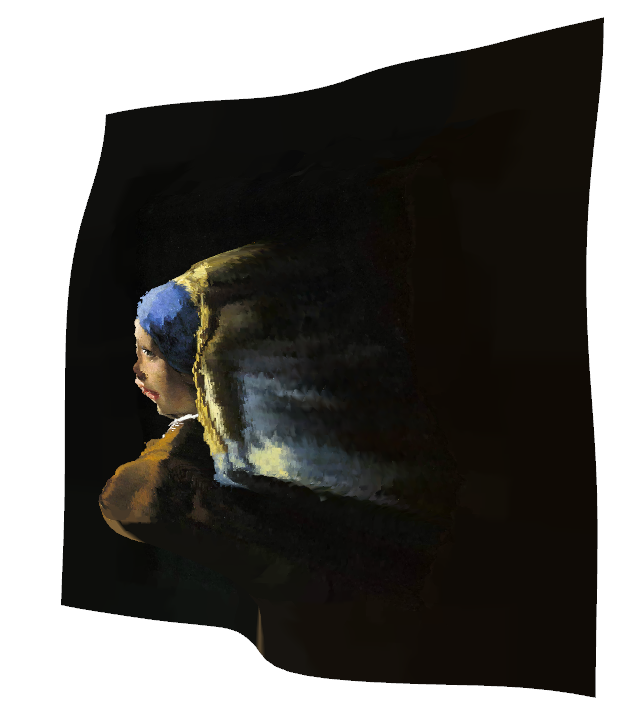}
         \caption{}
     \end{subfigure}
     \hfill
     \begin{subfigure}[b]{0.13\textwidth}
         \centering
         \includegraphics[width=\textwidth]{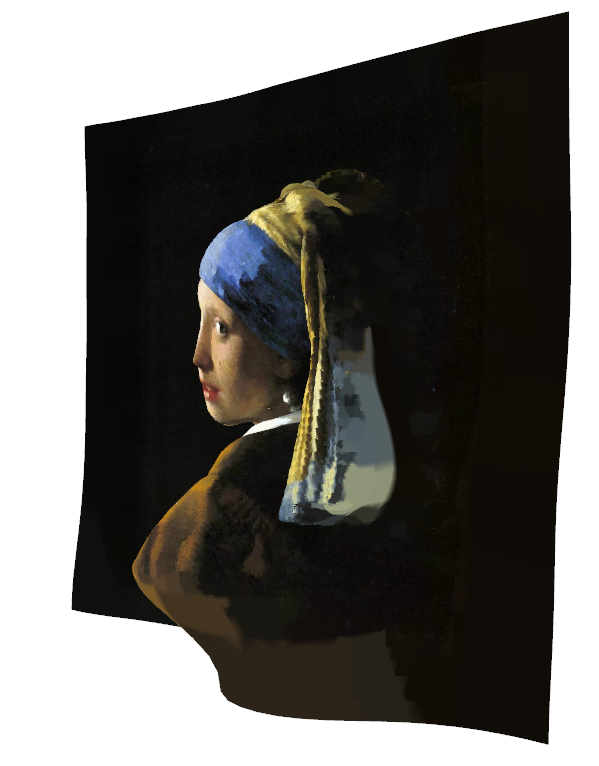}
         \caption{}
     \end{subfigure}
 
    \caption{Qualitative comparison of the 3D mesh reconstruction results, produced using Depth-Anything, GLPN and the proposed pipeline respectively.}
    \label{Fig:2}
\end{figure}

\subsection{Qualitative Analysis}
\label{subsec:qual_analysis}

In addition to the similarity-based evaluation, a qualitative analysis was performed to assess the visual plausibility and structural coherence of the 3D meshes generated by the proposed pipeline. This evaluation was conducted on a collection of images of artworks, primarily paintings, belonging to different artistic categories: portraits, landscapes, and geometric optical illusions. Portraits proved to be the category that yielded the most accurate results. In this case, the proposed pipeline produced highly detailed, well-proportioned meshes that were visually coherent with the original image. This can be attributed to the compositional care typically devoted to human subjects in paintings, where light and shadow effects, contrast, and volumetric rendering are emphasized by the artist to enhance facial features. These characteristics facilitate depth estimation, enabling the system to generate accurate and well-structured 3D reconstructions. Landscapes and optical illusions also produced satisfactory results, albeit with lower detail resolution. In landscape paintings in particular, visual depth and shape are often suggested through soft brushstrokes and gradually blending colors rather than sharp contours. While such visual styles are intuitively interpretable by the human eye, they may pose challenges for computational models that rely solely on visual patterns in the image. This difficulty is further exacerbated by the lack of multiple viewpoints, since the system operates from a single perspective of the subject, without access to additional angles as would be available, for instance, in photogrammetry. Despite these limitations, the results demonstrate the versatility of the proposed system in generating 3D models of artworks, regardless of the subject depicted. Although some variation in accuracy is observed depending on the pictorial detail, the pipeline proved capable of adapting effectively to a range of artistic styles. Moreover, the proposed solution outperformed the individual use of the two depth estimation models, benefiting from their combined strengths and the integration of targeted pre and post-processing steps. A qualitative comparison of the results produced by GLPN, Depth-Anything, and the pipeline presented in this study is shown in Fig. \ref{Fig:2}.

\section{Conclusion}
\label{sec:conclusion}
This paper presented the design and development of an augmented reality  system for museum environments, aimed at recognizing artworks and visualizing their 3D representations generated from single 2D images. The proposed pipeline combines monocular depth estimation models - GLPN and Depth-Anything - through a weighted strategy to balance global structure and fine detail extraction. Depth maps are transformed into point clouds and then into 3D meshes using the Poisson surface reconstruction method. These meshes are embedded in a Unity-based AR application to overlay virtual 3D models onto the corresponding real-world artworks. The study also identifies key challenges, particularly the difficulty of reconstructing geometrically accurate models from stylistic or abstract artworks using only a single image. Compared to multi-view approaches like photogrammetry, single-image reconstruction lacks depth cues, especially for subjects with minimal detail, such as landscapes. This limitation highlights the need for more tailored approaches in the context of fine art. Future directions include training depth estimation models on artwork-specific datasets and exploring modular integration of alternative or custom networks to enhance reconstruction quality.

\subsubsection{Acknowledgments.} This work was supported by ``Smart unmannEd AeRial vehiCles for Human likE monitoRing (SEARCHER)'' project of the Italian Ministry of Defence within the PNRM 2020 Program (PNRM a2020.231); ``EYE-FI.AI: going bEYond computEr vision paradigm using wi-FI signals in AI systems'' project of the Italian Ministry of Universities and Research (MUR) within the PRIN 2022 Program (CUP: B53D23012950001); ``Enhancing Robotics with Human Attention Mechanism via Brain-Computer Interfaces'' Sapienza University Research Projects (Grant Number: RM124190D66C576E).  
%
%
%
\bibliographystyle{splncs04}
\bibliography{references}

\begin{thebibliography}{10}
\providecommand{\url}[1]{\texttt{#1}}
\providecommand{\urlprefix}{URL }
\providecommand{\doi}[1]{https://doi.org/#1}

\bibitem{avola2022medicinal}
Avola, D., Cinque, L., Fagioli, A., Foresti, G.L., Marini, M.R., Mecca, A.,
  Pannone, D.: Medicinal boxes recognition on a deep transfer learning
  augmented reality mobile application. In: Sclaroff, S., Distante, C., Leo,
  M., Farinella, G.M., Tombari, F. (eds.) Image Analysis and Processing --
  ICIAP 2022. pp. 489--499. Springer International Publishing, Cham (2022)

\bibitem{avola2019slam}
Avola, D., Cinque, L., Fagioli, A., Foresti, G.L., Massaroni, C., Pannone, D.:
  Feature-based slam algorithm for small scale uav with nadir view. In: Ricci,
  E., Rota~Bul{\`o}, S., Snoek, C., Lanz, O., Messelodi, S., Sebe, N. (eds.)
  Image Analysis and Processing -- ICIAP 2019. pp. 457--467. Springer
  International Publishing, Cham (2019)

\bibitem{avola2020person}
Avola, D., Cinque, L., Fagioli, A., Foresti, G.L., Pannone, D., Piciarelli, C.:
  Bodyprint—a meta-feature based lstm hashing model for person
  re-identification. Sensors  \textbf{20}(18) (2020)

\bibitem{avola2019visual}
Avola, D., Cinque, L., Foresti, G.L., Pannone, D.: Visual cryptography for
  detecting hidden targets by small-scale robots. In: De~Marsico, M., di~Baja,
  G.S., Fred, A. (eds.) Pattern Recognition Applications and Methods. pp.
  186--201. Springer International Publishing, Cham (2019)

\bibitem{cherti2023reproducible}
Cherti, M., Beaumont, R., Wightman, R., Wortsman, M., Ilharco, G., Gordon, C.,
  Schuhmann, C., Schmidt, L., Jitsev, J.: Reproducible scaling laws for
  contrastive language-image learning. In: Proceedings of the IEEE/CVF
  Conference on Computer Vision and Pattern Recognition. pp. 2818--2829 (2023)

\bibitem{corno2025ar}
Corno, M., Franceschetti, L., Matteo~Savaresi, S.: Design of a cost effective
  spatial image registration system for augmented reality in vehicular
  applications. IEEE Transactions on Intelligent Transportation Systems
  \textbf{26}(3),  2967--2976 (2025)

\bibitem{Geiger2013IJRR}
Geiger, A., Lenz, P., Stiller, C., Urtasun, R.: Vision meets robotics: The
  kitti dataset. International Journal of Robotics Research (IJRR)  (2013)

\bibitem{ikeuchi2020light}
Ikeuchi, K., Matsushita, Y., Sagawa, R., Kawasaki, H., Mukaigawa, Y., Furukawa,
  R., Miyazaki, D.: Active Lighting and Its Application for Computer Vision.
  Springer Cham, 1 edn. (2020)

\bibitem{Kazhdan2006Poisson}
Kazhdan, M., Bolitho, M., Hoppe, H.: Poisson surface reconstruction. In:
  Sheffer, A., Polthier, K. (eds.) Symposium on Geometry Processing. The
  Eurographics Association (2006)

\bibitem{kerbl3Dgaussians}
Kerbl, B., Kopanas, G., Leimk{\"u}hler, T., Drettakis, G.: 3d gaussian
  splatting for real-time radiance field rendering. ACM Transactions on
  Graphics  \textbf{42}(4) (2023)

\bibitem{Kim2022GLPN}
Kim, D., Ka, W., Ahn, P., Joo, D., Chun, S., Kim, J.: Global-local path
  networks for monocular depth estimation with vertical cutdepth. CoRR
  \textbf{abs/2201.07436} (2022), \url{https://arxiv.org/abs/2201.07436}

\bibitem{kim2023clip}
Kim, K., Kim, M.J., Kim, H., Park, S., Paik, J.: Person re-identification
  method using text description through clip. In: 2023 International Conference
  on Electronics, Information, and Communication (ICEIC). pp.~1--4 (2023)

\bibitem{oquab2024dinov2learningrobustvisual}
Oquab, M., Darcet, T., Moutakanni, T., Vo, H., Szafraniec, M., Khalidov, V.,
  Fernandez, P., Haziza, D., Massa, F., El-Nouby, A., Assran, M., Ballas, N.,
  Galuba, W., Howes, R., Huang, P.Y., Li, S.W., Misra, I., Rabbat, M., Sharma,
  V., Synnaeve, G., Xu, H., Jegou, H., Mairal, J., Labatut, P., Joulin, A.,
  Bojanowski, P.: Dinov2: Learning robust visual features without supervision
  (2024), \url{https://arxiv.org/abs/2304.07193}

\bibitem{radford2021learningtransferablevisualmodels}
Radford, A., Kim, J.W., Hallacy, C., Ramesh, A., Goh, G., Agarwal, S., Sastry,
  G., Askell, A., Mishkin, P., Clark, J., Krueger, G., Sutskever, I.: Learning
  transferable visual models from natural language supervision (2021),
  \url{https://arxiv.org/abs/2103.00020}

\bibitem{ranftl2021midas}
Ranftl, R., Bochkovskiy, A., Koltun, V.: Vision transformers for dense
  prediction. In: 2021 IEEE/CVF International Conference on Computer Vision
  (ICCV). pp. 12159--12168 (2021)

\bibitem{remondino2006image}
Remondino, F., El-Hakim, S.: Image-based 3d modelling: A review. The
  Photogrammetric Record  \textbf{21}(115),  269--291 (2006)

\bibitem{Silberman:ECCV12}
Silberman, N., Hoiem, D., Kohli, P., Fergus, R.: Indoor segmentation and
  support inference from rgbd images. In: Fitzgibbon, A., Lazebnik, S., Perona,
  P., Sato, Y., Schmid, C. (eds.) Computer Vision -- ECCV 2012. pp. 746--760.
  Springer Berlin Heidelberg, Berlin, Heidelberg (2012)

\bibitem{artgan2018}
Tan, W.R., Chan, C.S., Aguirre, H., Tanaka, K.: Improved artgan for conditional
  synthesis of natural image and artwork. IEEE Transactions on Image Processing
   \textbf{28}(1),  394--409 (2019)

\bibitem{tripoSR2024}
Tochilkin, D., Pankratz, D., Liu, Z., Huang, Z., , Letts, A., Li, Y., Liang,
  D., Laforte, C., Jampani, V., Cao, Y.P.: Triposr: Fast 3d object
  reconstruction from a single image. arXiv preprint arXiv:2403.02151  (2024)

\bibitem{tron2016sfm}
Tron, R., Zhou, X., Daniilidis, K.: A survey on rotation optimization in
  structure from motion. In: 2016 IEEE Conference on Computer Vision and
  Pattern Recognition Workshops (CVPRW). pp. 1032--1040 (2016)

\bibitem{depthanything}
Yang, L., Kang, B., Huang, Z., Xu, X., Feng, J., Zhao, H.: Depth anything:
  Unleashing the power of large-scale unlabeled data. In: CVPR (2024)

\end{thebibliography}
\end{document}